\documentclass{IOS-Book-Article}

\usepackage{mathptmx}
\usepackage{cite}
\usepackage{amsmath,amssymb,amsfonts}
\usepackage{algorithmic}
\usepackage{graphicx}
\usepackage{textcomp}
\usepackage{xcolor}

\usepackage{graphicx}
\usepackage{amsmath,amssymb,amsfonts}
\usepackage{algorithmic}
\usepackage{graphicx}
\usepackage{import}
\usepackage{hyperref}
\usepackage{amsmath}
\usepackage{algorithm}  
\usepackage{algorithmic}

\usepackage{booktabs}
\usepackage{multirow}

\usepackage{times}
\usepackage{soul}
\usepackage{url}
\usepackage[utf8]{inputenc}
\usepackage{amsthm}
\usepackage{soul}\setuldepth{article}
\def\hb{\hbox to 11.5 cm{}}

\begin{document}

\pagestyle{headings}
\def\thepage{}
\begin{frontmatter}              

\title{Recognizing Entity Types via Properties}


\author{\fnms{Daqian} \snm{Shi}
}and 
\author{\fnms{Fausto} \snm{Giunchiglia}
}
\address{Department of Information Engineering and Computer Science \\(DISI), University of Trento, Italy}

\begin{abstract}
The mainstream approach to the development of ontologies is merging ontologies encoding different information, where one of the major difficulties is that the heterogeneity motivates the ontology merging but also limits high-quality merging performance. Thus, the entity type (etype) recognition task is proposed to deal with such heterogeneity, aiming to infer the class of entities and etypes by exploiting the information encoded in ontologies. In this paper, we introduce a \textit{property-based} approach that allows recognizing etypes on the basis of the properties used to define them. From an epistemological point of view, it is in fact properties that characterize entities and etypes, and this definition is independent of the specific labels and hierarchical schemas used to define them. The main contribution consists of a set of property-based metrics for measuring the contextual similarity between etypes and entities, and a machine learning-based etype recognition algorithm exploiting the proposed similarity metrics. Compared with the state-of-the-art, the experimental results show the validity of the similarity metrics and the superiority of the proposed etype recognition algorithm.
\end{abstract}

\end{frontmatter}
\markboth{April 2022\hb}{April 2022\hb}

\section{Introduction}
The web has enabled the generation and sharability of a virtually unbound quantity of data, where ontology is by far one of the most widely used representation models \cite{chen2019identifying}. The main motivation is the inherent flexibility of the ontology model, for instance, not all entities need to have the same properties and an ontology can be evolved whenever new data become available. This flexibility is key to extending ontologies by merging diverse ontologies and data resources, thus exploiting the available information about different entities. However, one of the major difficulties is that the heterogeneity motivates ontology merging but also limits high-quality merging performance. Thus, the entity type (etype) recognition task is proposed to deal with such heterogeneity, aiming to infer the class of entities and etypes by exploiting the information encoded in ontologies. According to the application scenarios, we consider two cases, where schema-level etype recognition \cite{algergawy2018results} aims at aligning a set of candidate etypes to a set of reference etypes, and instance-level etype recognition \cite{portisch2021background} aims to predict the etype of given entities. 

Existing etype recognition methods mainly exploit lexical-based \cite{ sun2015comparative} and structure-based \cite{faria2013agreementmakerlight} techniques. Both techniques enforce etype label matching as a prerequisite, which utilizes diverse lexical similarity metrics and synonym analysis to align etype labels. Such methods are limited when applied in practice since the same concept can be labeled differently by ontologies \cite{sleeman2015entity}, e.g., an eagle can be labeled as \textit{Bird} in a general-purpose ontology and \textit{Eagle} in a domain-specific ontology. In turn, the same label may present different concepts in heterogeneous ontologies, which will also lead to wrong recognition results. Structure-based methods consider the hierarchy as an additional input to drive the label matching, e.g., matching on super-classes. However, these methods may also mislead the conclusions as properties assigned to an etype in the hierarchy are cumulative and do not depend on the order by which they are assigned \cite{giunchiglia2020entity}. In addition, the difference in taxonomy between ontologies will increase the impact of such issues, e.g., the super-class of etype \textit{Eagle} can be \textit{Animal} in one ontology and \textit{Bird} in another. 

As a solution to the above problems, the key intuition behind our approach, is to recognize etypes based on the \textit{properties} which define them.  From an epistemological point of view, it is in fact properties that implicitly define the etype, and this definition is independent of the specific label used to name a concept, and of the specific hierarchical schema of the ontology used to define the etype. This allows us to exploit the fact that etypes are organized in hierarchies, where lower etypes inherit properties from upper etypes and where the entities populating an etype also populate all the upper etypes. Thus, based on the above intuitions, we propose a general property-based algorithm for etype recognition problem.  We introduce a formalization strategy where the ontology and its inner mappings are organized by formal concept analysis (FCA) \cite{ganter2012formal}. We present three \textit{property-based} metrics to measure the contextual similarity between etypes and entities, where the metrics characterize the role that properties have in the definition of given etypes from different aspects. They capture the main idea that the number of aligned properties affects the contextual similarity between concepts. 

Overall, the main contributions of this paper are as follows:
\begin{itemize}
    \item We design an ontology formalization strategy and a novel set of metrics for measuring contextual similarity across reference etypes and candidates.
    \item We propose a machine learning (ML)-based algorithm that implements etype recognition as a classification task, via the similarity metrics mentioned above.
    \item We compare our method with state-of-the-art on several benchmarks. The experimental results show the validity of the similarity metrics and the superiority of the proposed etype recognition algorithm.
\end{itemize}

The rest of the paper is organized as follows. Section 2 discusses the intuition for exploiting properties and demonstrates how to formalize an ontology into an FCA context. Section 3 introduces three specificity measurements and their corresponding property-based etype similarity metrics. In section 4, we describe the proposed ML-based etype recognition algorithm, and we present the experimental setups and results in section 5. Finally, we present the related work in section 6 and conclude the paper in section 7.

\section{Schemas, Etypes and Entities}
\label{sec3}

To clearly present the task we discussed, we define the schema of an ontology and its inner relations as $Sch = \langle C, P, R  \rangle$, where $C = \{C_1,…,C_n\}$ being etypes, $P = \{p_1,…,p_m\}$ being the set of properties, $R = \{\langle C_i,T(C_i) \rangle|C_i \in C \}$ being the set of correspondences between etypes and properties, and function $T(C_i)$ returns properties associated with $C_i$. Let us also define $I = \{I_1,…,I_l\}$ as a set of entities, where each entity $I_i$ is associated with specific etype $C_i$, and $t(I_i)$ returns a set of associations between entities and properties. We consider that the property $p_i$ is used to describe an etype $C_i$ or an entity $I_i$ when the property belongs to set $T(C_i)$ or $t(I_i)$, respectively. Thus, given a reference ontology $Ont_{ref}$, the goal of etype recognition is to match the etype $C_{ref}$ with $C_{cand}$ and predict the etype $C_{ref}$ of entities $I_{cand}$ from the candidate ontology $Ont_{cand}$, where $C_{cand}, I_{cand} \in Ont_{cand}$, $C_{ref} \in Ont_{ref}$.


\begin{table*}[!t]
\centering
\caption{Shared Properties of etype \textit{Person} across different ontologies.}
  \label{PropertyExample}

\begin{tabular}{@{}lcccl@{}}
\toprule
\textbf{Contexts}      & & \textbf{Tot.} &  & \textbf{Shared Properties} \\  \midrule
OpenCyc \& DBpedia     & & 39  & & \textit{birth, education, title, activity, ethnicity, employer, status...} \\
OpenCyc \& Schema.org  & & 21  & & \textit{contact, suffix, tax, job, children, works, worth, gender, net...} \\
DBpedia \& FreeBase    & & 33  & & \textit{title, number, related, birth, parent, work, name...} \\ 
DBpedia \& Schema.org  & & 22  & & \textit{death, sibling, point, member, nationality, award, parents...} \\ 
\bottomrule                    
\end{tabular}
\end{table*}

\begin{figure*}[!t]
\setlength{\abovecaptionskip}{0pt}%
\setlength{\belowcaptionskip}{0pt}%
	\centering
	\includegraphics[width=0.99\linewidth]{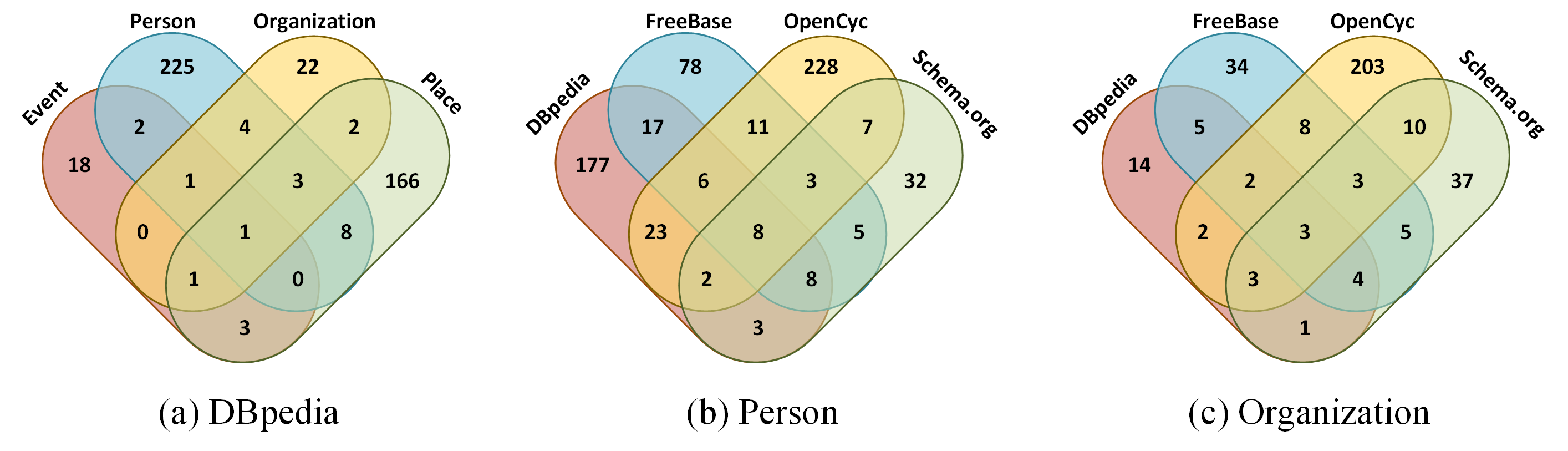}
	\caption{The shareability of properties that occurs within and across ontologies. 
	\label{fig:1}}
\end{figure*}

The intuition of utilizing properties for etype recognition comes from the property being one of the most basic and critical elements for implicitly defining etypes \cite{giunchiglia2021property}. For each schema, etypes play the role of categorization, and properties aim to draw sharp lines so that each entity in the domain falls determinedly either in or out of each etype \cite{giunchiglia2019knowledge}. Meanwhile, we have  following observations when comparing properties across different ontologies: (1) In a specific ontology, each etype is described by a set of properties, whereas most of the properties are distinguishable according to the belonging etypes and a small number of properties are shared across different etypes; (2) Same or similar properties are shared across different ontologies for describing the same concepts.
To visually present these observations, we present shared properties of etype \textit{Person} across different ontologies in Table \ref{PropertyExample}, e.g., \textit{birth} and \textit{education} are applied in both OpenCyc and DBpedia. All these examples present features of properties, namely, the unity for describing the same concept and the diversity for distinguishing different concepts. For instance, a \textit{Person} can be distinguished from a \textit{Place} by the property \textit{birth}, which is a crucial step to identify the etypes. Meanwhile, we also introduce a special type of \textit{Venn graphs}, namely \textit{knowledge lotuses}, to prove our observations by representing the shareability of properties that occurs within and across ontologies. Knowledge lotuses provide a synthetic view of how different ontologies or etypes overlap in properties \cite{giunchiglia2020entity}. Figure \ref{fig:1} shows several examples, where we assume that we have four contexts built from (parts of) the four biggest knowledge bases, i.e., OpenCyc, DBpedia, Schema.org and FreeBase. Each value in a lotus refers to the number of shared properties. Thus, we find that it is important to exploit and measure properties for better etype recognition.

\begin{figure*}[!t]
	\centering
	\includegraphics[width=0.99\linewidth]{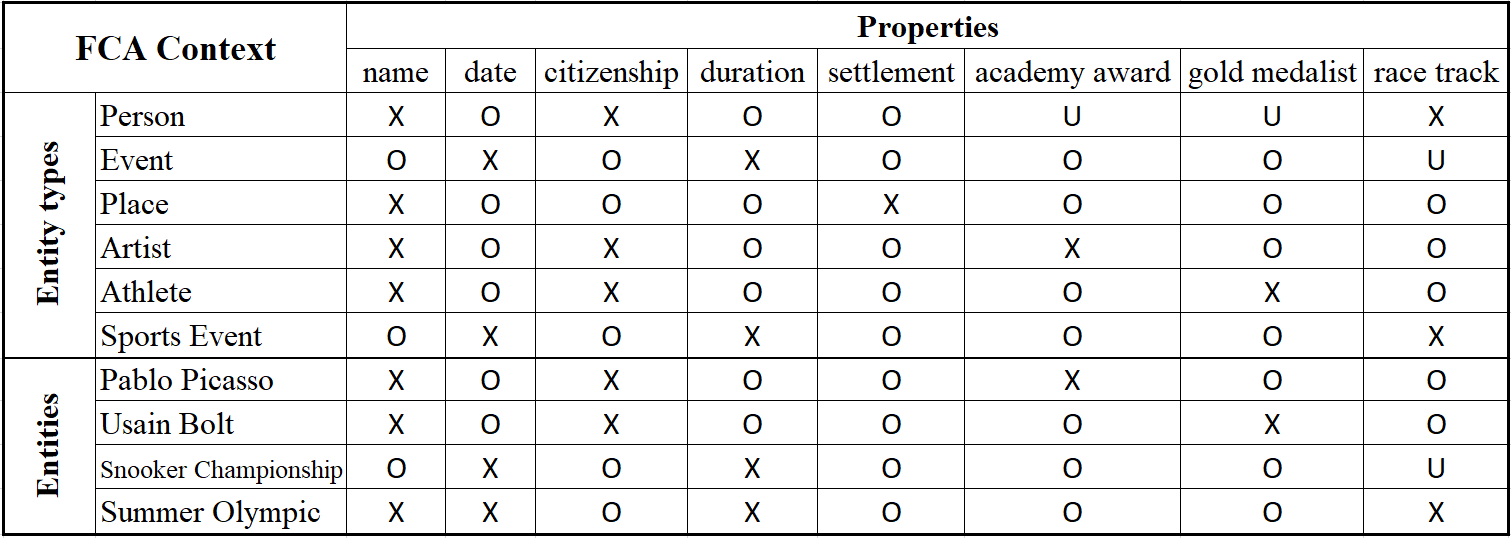}
	\caption{An example of formalizing ontology into FCA contexts 
	\label{fig:2}}
\end{figure*}

To utilize the property information, we formalize the relation between properties and etypes/entities as \textit{associations}. Thus, two cases are considered. At the schema level, the ontology schema will be flattened into a set of triples, where each triple encodes information about \textit{etype-property} associations, e.g., triple “organization-domain-LocatedIn” encodes the “organization-LocatedIn” association. Instance-level cases generally define triples as “entity-property-entity", where two associations are encoded. For instance, triple “EiffelTower-LocatedIn-Paris” encodes “EiffelTower-LocatedIn" and “Paris-LocatedIn". To generate our FCA contexts, we have the following settings: (1) Compared to general FCAs, we consider both etypes $C$ and entities $I$, where we associate an entity with its properties $t(I_i)$, and also an etype with its properties $T(C_i)$; (2) We introduce the notion of \textit{undefined} to describe an additional relation between etypes and properties. As an example, we generate an FCA context whose information is extracted from DBpedia \cite{auer2007dbpedia}, as shown in Figure \ref{fig:2}. We adopt the following conventions. The value box with a cross represents the fact the property is associated with the etype, e.g., \textit{citizenship} is associated with \textit{Person}. The value box with a circle means the property is unassociated with the etype, e.g., \textit{date} as for \textit{Person}. The value “U” (for undefined) represents the fact that the property is unassociated with the etype but associated with one of its sub-classes. The intuition is that the property might or might not be used to describe the current etype, e.g., \textit{academy award} is used to describe \textit{Artist} and it might be used to describe \textit{Person}. Similar to etypes, we can also find formalized entities and their properties. Note that these entities are selected from the ontology with a hierarchical schema, thus, they can inherit the \textit{unassociated properties} from their etypes, e.g., as an \textit{Athlete}, \textit{Usain Bolt} does not have property \textit{duration}. 

As a result, we encode the above-mentioned three correlations as the parameter $w_E(p)$. Considering the correlation of “associated with” is positive for a property-based description, the correlation of “unassociated with” is negative and the correlation of “undefined” is neutral, we define the parameter as:
\begin{equation}
w_E(p) = \left\{
\begin{tabular}{ll}
1,  & if $p \in prop(E)$\\
0,  & if $p \notin prop(E) \land p \in prop(E.subclass)$ \\
-1, & if $p \notin prop(E) \land p \notin prop(E.subclass)$ 
\end{tabular}\right.
\label{equ:1}
\end{equation}
where we suppose $E$ is a general concept that unifies the definitions of etypes $C$ and entities $I$ in an ontology, $p$ is the target property, $E.subclass$ refers to the sub-classes of the etype $E$, and $prop(E)$ refers to the properties associated with $E$. Need to notice a special case that \textit{undefined properties} also exists where a specific entity misses the inherited property, such as \textit{race track} is used to describe \textit{Sports events} but missed for its entity \textit{Snooker Championship}, which will also make $w_E(p)=0$. Thus, the crosses, “U”s and circles in Figure \ref{fig:2} are set to 1, 0 and -1, respectively.

\section{Property-based Similarity Metrics}
\label{sec4}
The property-based similarity metrics we proposed in this paper are inspired by \cite{fumagalli2021ranking} which introduces the importance of properties for describing etypes. The key intuition is that properties at different levels of specificity contribute differently during etype recognition. e.g., a more specific property provides more information that allows identifying an etype \cite{giunchiglia2007formalizing}. This results in the definition of \textit{horizontal specificity}, \textit{vertical specificity}, and \textit{informational specificity} and of their corresponding similarity metrics.

\subsection{Horizontal specificity}

When measuring the specificity of a property, a possible idea is to horizontally compare the number of etypes that are described by a specific property, i.e., the shareability of the property \cite{giunchiglia2020entity}. If a property is used to describe diverse etypes, it means that the property is not highly characterizing its associated etypes. For instance, the property \textit{name} is used to describe \textit{Person}, \textit{Place}, and \textit{Athlete}, where \textit{name} is a common property that appears in different contexts. \textit{Education} is a highly specific property since it is associated only with the etype \textit{Person}. Based on this intuition, we consider the specificity of a property as related to its shareability. Thus, we have  horizontal specificity ($HS$) to measure the number of etypes that are associated with the target property, as: 
\vspace{-5pt}
\begin{equation}
HS_{Ont}(E,p)= w_E(p) * {e^{\lambda(1-|K_v|)}} \in [-1, 1]
\label{equ:2}
\end{equation}
where $p$ is the input property, $E$ is the input etype/entity and $K_v$ is the set of etypes described by $p$ in a specific ontology $Ont$; $|K_v|$ is the cardinality of $K_v$, thus $|K_v| \ge 1$; $e$ denotes the natural mathematical constant \cite{finch2003mathematical}; $\lambda$ represents a constraint factor. 

\subsection{Vertical specificity}

Etypes are organized into classification hierarchies such that the upper-layer etypes represent more general concepts, whereas the lower-layer etypes represent more concrete concepts \cite{giunchiglia2007formalizing}. 
Thus, properties of lower-layer etypes are more specific since they are used to describe specific concepts. We assume that lower-layer properties will contribute more to the etype identification since they are more specific. For instance, as a lower-layer etype, \textit{Artist} can be identified by the property \textit{academy award} but not by the general property \textit{name}. Based on this intuition, we propose $VS$ for capturing the vertical specificity, where $layer(E)$ refers to the layer of the inheritance hierarchy to define $E$. 
\vspace{-5pt}
\begin{equation}
VS_{Ont}(E,p)= w_E(p) * \min_{E \in K_v} layer(E) \in [-1, 1]
\label{equ:3}
\end{equation}

\subsection{Informational specificity}

Based on the observation that the property specificity should also take into account the effect of the number of entities, we introduce the notion of informational specificity $IS$ which will change with the number of entities populating it. The intuition is that specificity will decrease when the entity counting increases. For instance, the $IS$ of \textit{academy award} decreases when there are increasing entities of \textit{Artist}. 
The definition of informational specificity is inspired by Kullback–Leibler divergence theory \cite{van2014renyi}, which is introduced to measure the difference between two sample distributions $Y$ and $\hat{Y}$. 
In the definition of informational specificity, we need to exploit some notions from information theory. We define informational entropy as: 
\vspace{-5pt}
\begin{equation}
H(K_v)= \frac{- \sum_{i=1}^{|K_v|} F(E_i)\log\frac{F(E_i)}{|K_v|}}{|K_v|}
\label{equ:4}
\end{equation}
where $K_v$ refers to any subset of $K$ in an ontology; $H(\cdot)$ represents the informational entropy of an etype set; $E_i$ is a specific etype in set $K_v$; $F(E_i)$ refers to the number of samples of etype $E_i$. 
After calculating informational entropy, the informational specificity $IS$ is defined as:
\vspace{-5pt}
\begin{equation}
IS_{Ont}(E,p)= w_E(p) * (H(K) - \sum \frac{|K_v|}{|K|} H(K_v)) \in [-1, 1]
\label{equ:5}
\end{equation}
where we weight each $H(K_v)$ by the proportion of $|K_v|$ to $|K|$. Being subtracted by the overall informational entropy $H(K)$, $IS$ presents the importance of the property $p$ for describing the given etype set $K$. 

\subsection{Similarity metrics}
We have modeled the specificity of properties, which represent their weights for describing ontologies from different aspects. Then, we define three property-based similarities based on the corresponding specificity. Given a reference ontology $Ont_A$ and a candidate ontology $Ont_B$, we define a function for calculating different similarities between etypes and entities from $Ont_A$ and $Ont_B$ based on their corresponding specificity: 
\begin{equation}
\small
Sim(E_a,E_b) =\frac{1}{2}\sum_{i=1}^{k} \left ( \frac{SPC_{A}(E_a,p_i)}{|prop(E_a)|} + \frac{SPC_{B}(E_b,p_i)}{|prop(E_b)|} \right) \in [0, 1]
\label{equ:6}
\end{equation}

\noindent where we take $E_a$, $E_b$ as the etypes/entities from $Ont_A$ and $Ont_B$ respectively, thus $E_a \in Ont_A$ and $E_a \in Ont_B$; $prop(E)$ refers to the properties associated with the specific etype and $|prop(E)|$ is the number of properties in $prop(E)$; $SPC_{ETG}(\cdot)$ represents the specificity measurements we defined above, $SPC(\cdot) = \{HS(\cdot),VS(\cdot),IS(\cdot) \}$, thus $SPC_{A}(E_a,p_i)$ and $SPC_{B}(E_b,p_i)$ refer to the specificity of the aligned property $p_i$ from two ontologies; $k$ is the number of aligned properties which are associated with both etype $E_a$ and $E_b$. As a result, we obtain three similarity metrics which are horizontal similarity $Sim_H$, vertical similarity $Sim_V$, and informational similarity $Sim_I$. 
Notice that each similarity metric is symmetric, more specifically, $Sim(E_a,E_b) = Sim(E_b,E_a)$. Note also that we apply z-score normalization \cite{patro2015normalization} to similarity metrics at the end of calculations, which makes the range of $Sim_H, Sim_V ,Sim_I$ between $0$ to $1$.

\section{Etype Recognition Algorithm}
\label{sec5}
\begin{figure*}[!t]
	\centering	\includegraphics[width=0.6\linewidth]{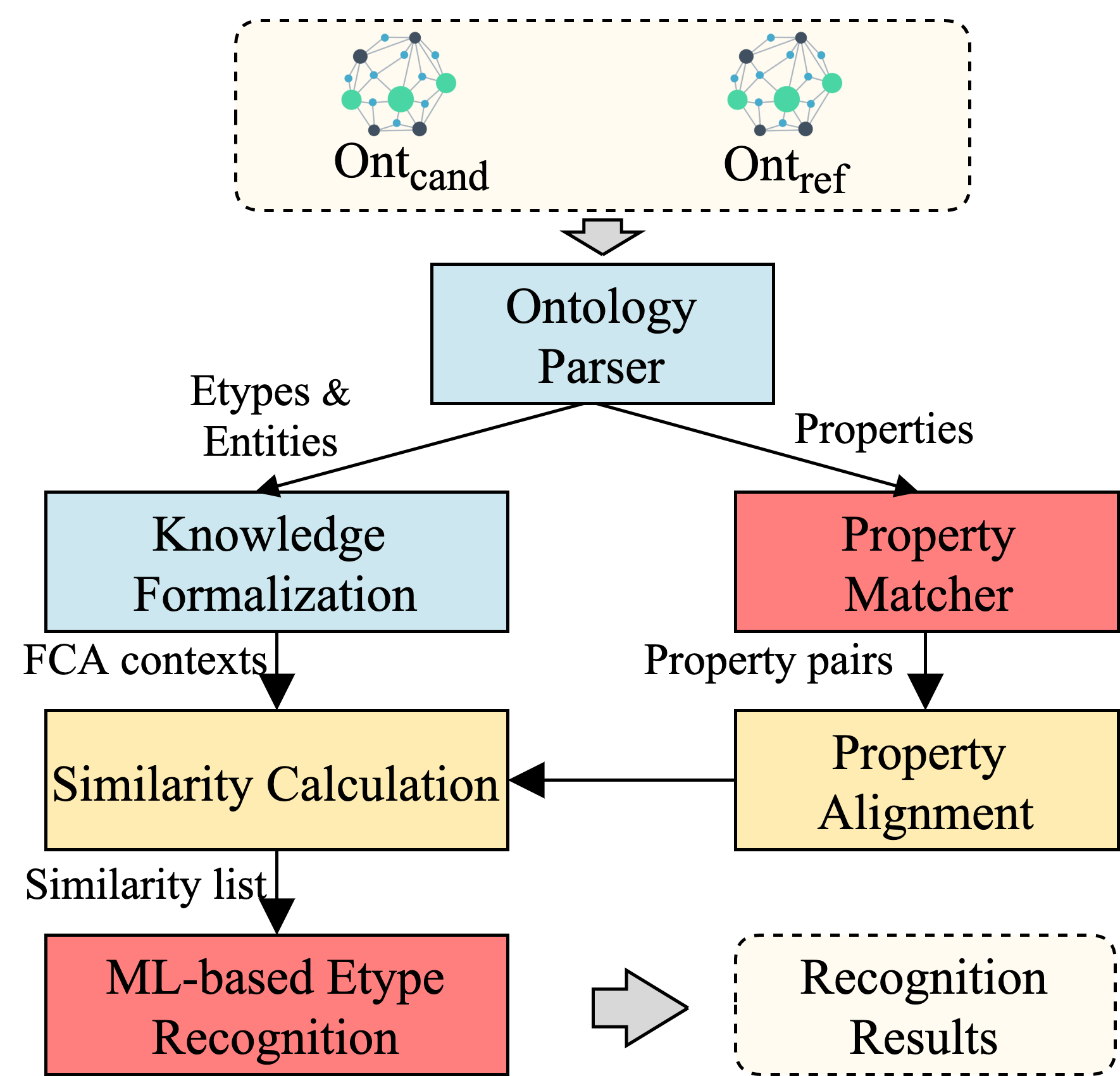}
	\caption{Overall framework of our etype recognition algorithm\label{fig:3}}
\end{figure*}

In order to predict the etype of unknown concepts from candidate ontologies, we propose an etype recognition algorithm that exploits the property-based similarity metrics defined above where, as Figure \ref{fig:3} shows, different modules are marked in different colours. 
The ontology parser aims to parse the input ontology as a structured set of etypes, entities, and properties. The ontology will be flattened into a set of triples, where properties and concepts will be extracted from triples and then formalized as an FCA context. These two modules are marked in blue since they are pre-processing modules. All properties are collected and sent to the natural language processing (NLP)-based property matcher. Different labels of properties may express the same meaning since many of them are minor variations of the same label. Thus, an NLP pipeline is designed to normalize all input properties, where phrase segmentation, lemmatization and stop-word removal are introduced for better normalization performance. Then, string-based and language-based similarity metrics are exploited for matching the properties with normalized labels \cite{bella2017language,bella2016domain}. Thus, the property matcher will output the list of aligned properties. In the next phase, we generate the property-based etype similarities $Sim_H$,  $Sim_V$ and $Sim_I$ by FCA contexts and aligned properties. According to the Function (\ref{equ:6}), three similarity values will be generated for each etype pair, which will then be passed to the etype recognition module for final results. Note that ML-based modules (matchers) are marked in red.

We develop an ML-based method that implements etype recognition into a \textit{binary classification task}. The main idea is to predict if two incoming concepts are aligned with each other. For applying this method, a list of candidate pairs is prepared by pairing etypes/entities from candidate ontology $Ont_{cand}$ with etypes from reference ontology $Ont_{ref}$. Etypes $C_{ref} \in Ont_{ref}$ will be outputted as the final recognition results when the result of classification is ``aligned". For increasing recognition performance, we also exploit string-based and language-based similarity metrics, along with property-based similarity metrics $Sim_H$,  $Sim_V$, and $Sim_I$ for training and predicting.
The \textit{Property matcher} aims to align properties between ontologies, using the same prediction strategy as for etype recognition module. Considering that it is inevitable to have misaligned candidate pairs, we discuss the following solutions to reduce the effect of misaligned properties for better applying etype recognition. Firstly, the use of similarity metrics allows recognizing etypes by soft aligning, even if there are a few properties not aligned. This will increase the robustness of our etype recognition approach. Secondly, we define a specific category of properties named “undefined” properties (neutral), which will not affect the model training and reduce the additional interference since misaligned “undefined” properties will not be used for similarity calculation. Finally, by learning from the practical data from different resources, ML models will propose a learnable strategy rather than a fixed threshold for determining alignments, which will maximize the use of existing aligned properties and minimize the effect of misaligned properties.

\section{Evaluation}
\label{sec6}

\subsection{Experimental setup}

\noindent \textbf{Dataset selection.} 
We conduct experiments on several real-world datasets used for ontology matching and entity typing aiming at comprehensively evaluating our proposed algorithm. Our approach focuses on ontologies that contain etypes associated with a fair number of properties. For the evaluation of schema-level recognition cases (etype-etype alignments), we involve CONF (ConfTrack $ra1$\footnote{https://owl.vse.cz/ontofarm/} version) \cite{zamazal2017ten} and BIBLIO (BiblioTrack) \cite{euzenat2010results} from Ontology Alignment Evaluation Initiative\footnote{http://oaei.ontologymatching.org/2021/} (OAEI) tracks which are the main references for most of the schema matching methods, where dataset CONF contains 21 annotated reference ontology pairs, and BIBLIO contains 6 reference ontology pairs. For validating our algorithm on instance-level etype recognition cases (entity-etype alignments), we build two datasets, namely EnType$_{Self}$ and EnType$_{Gen}$, since there is no publicly released dataset for such etype recognition tasks that involves more than one ontology. We exploit the public dataset DBpedia630k \cite{zhang2015character} and DBpedia infobox\footnote{http://wikidata.dbpedia.org/services-resources/ontology} as the reference ontology for providing reference etypes. Because DBpedia is a general-purpose ontology that contains common etypes in the real world, where sufficient properties are applied for describing these etypes. Then we select candidate entities from DBpedia, SUMO\footnote{https://www.ontologyportal.org/} and several domain-specific datasets \cite{tang2008arnetminer}. The entities we selected mainly according to common etypes, more specifically, \textit{Person, Place, Event, Organization} and their sub-classes. Finally, we obtain 20,000 entity-etype candidate pairs, where 6,000 from DBpedia (EnType$_{Self}$) and 14,000 from the remaining resources (EnType$_{Gen}$). Notice that we randomly select training samples for all datasets we applied in the experiments to keep fair comparisons, where each dataset is divided into a training set with 50\%, test set with 30\%, and validation set with 20\%. Also notice that there are no shared entities between the train, test, and validation sets for all the datasets. 

\vspace{0.1cm}
\noindent \textbf{Feature selection.}
Our etype recognition algorithm applies a general binary classification strategy, which is independent of the specific ML model. Thus, the data label of positive and negative samples refers to if the candidate pair is matched. The data consists of three kinds of features, which are property-based similarity metrics ($Sim_H$, $Sim_V$ and $Sim_I$), string-based similarity metrics (N-gram \cite{euzenat2007ontology}, Longest common sub-sequence \cite{euzenat2007ontology}, Levenshtein distance \cite{yujian2007normalized}) and language-based similarity metrics (Wu and Palmer similarity \cite{palmer1994verb} and Word2Vec \cite{church2017word2vec}). Besides property-based similarity metrics, some of the beneficial string-based and language-based metrics are selected as additional metrics for achieving better recognition performance. These similarity metrics aim to measure different aspects of the relevance between the reference etypes and candidates. Since all the above-mentioned similarity metrics are symmetric, the order of etype/entity in the candidate pair will not affect the final results. Moreover, we apply only property-based similarity metrics ($Sim_H$, $Sim_V$ and $Sim_I$) for instance-level etype recognition. Because the label of the candidate entity is commonly not relevant to its etype (e.g., entity \textit{Bob} and its etype \textit{Person}).


\vspace{0.1cm}
\noindent \textbf{Evaluation metrics.}
In our experiment, we exploit widely-used evaluation metrics, including Macro-averaged (Ma-F1) and Micro-averaged $F_1$-measure (Mi-F1), to compare our algorithm with state-of-the-art methods. We form the etype recognition candidates as pairs, where each pair consists of a reference etype and a candidate etype/entity. F1 scores Ma-F1 and Mi-F1 are calculated from the basic metrics precision and recall which are respectively referring to the fraction of correctly identified pairs among all identified pairs and the fraction of correctly identified pairs among all the ground-truth pairs. We consider F1 scores to be the most relevant metrics for our evaluation since they reflect both recall and precision. Notice that Ma-F1 focuses on evaluating the performance of each class and Mi-F1 evaluates the overall performance of all samples.

\subsection{Experimental results}

\subsubsection{Qualitative analysis}
\begin{table}[!t]
\centering
\setlength{\abovecaptionskip}{0pt}    
\setlength{\belowcaptionskip}{1pt}
\caption{Representative samples of property-based similarity $Sim_V$, $Sim_H$ and $Sim_I$ between candidate etype-etype (left) and entity-etype (right) entity pairs.}
  \label{tab2}
\resizebox{1\columnwidth}{!}{
\begin{tabular}{|c|c|c|c|c|c||c|c|c|c|c|c|}
\hline

 $C_{cand}$ & $C_{ref}$          & $Sim_V$ & $Sim_H$ & $Sim_I$ & M & $I_{cand}$    &   $C_{ref}$     & $Sim_V$ & $Sim_H$ & $Sim_I$ & M\\ \hline
    Contribution  &   Paper        & 1      & 0.853  & 0.730 & $\times$ &  MiltHinton     &    Person    & 1      & 0.787  & 0.873 & $\times$ \\ \hline
Topic &       SubjectArea         & 0.756  & 0.740  & 0.857 & $\times$ &  Jadakiss     &      Person     & 1      & 0.645  & 0.305 & $\times$ \\ \hline
 Topic &       Author          & 0.198  & 0.353  & 0.018 & & Boston      &       Person      & 0.264  & 0.173  & 0.041 & \\ \hline
 Poster &       Meta-Review         & 0      & 0.312  & 0.262 & &   Boston     &      Place       & 0.720  & 1      & 0.433 & $\times$ \\ \hline
 Chair   &      Chairman           & 1      & 0.559  & 0.554 & $\times$ &   Jadakiss  &    Place          & 0.128  & 0.093  & 0.148 & \\ \hline

  Publisher  &     Chairman        & 0      & 0.07   & 0.195 & &  MiltHinton &   Organization     & 0.070  & 0.022  & 0.092 & \\ \hline
\end{tabular}
}
\end{table}

Table \ref{tab2} provides representative examples to show the etype similarity metrics between candidate etype-etype pairs from ConfTrack and candidate entity-etype pairs from EnType$_{Gen}$, respectively. Value box “M" demonstrates if two etypes are referring to the same concept, where $\times$ refers to a positive answer. We find that the value of our property-based similarity metrics indeed captures the contextual similarity between relevant etypes, where aligned etypes output higher values (e.g., \textit{paper-contribution}), in turn, non-aligned etypes return lower values (e.g., \textit{person-document}). With a broad observation of the metric values, we consider that our proposed property-based similarity metrics $Sim_H$, $Sim_V$ and $Sim_I$ are valid for both cases of etype recognition.

\subsubsection{Quantitative evaluation}

Our etype recognition method is named ETR$_{XGBoost}$ and ETR$_{ANN}$, where $XGBoost$ and $ANN$ refer to machine learning models XGBoost classifier \cite{chen2016xgboost} and artificial neural network (ANN) classifier \cite{nath2021automated}, respectively. We apply two models in our algorithm as a kind of ablation study to comprehensively evaluate the validity of our proposed similar metrics on the etype recognition task. We have compared our work with state-of-the-art matching methods, including previous OAEI evaluation campaigns (FCAMap \cite{chen2019identifying}, AML \cite{faria2013agreementmakerlight}, LogMap \cite{jimenez2011logmap}), ML-based ontology matching methods (Bento et. al. \cite{bento2020ontology}, Nkisi-Orji et. al. \cite{nkisi2018ontology}), deep learning-based entity typing methods (ConnectE \cite{zhao2020connecting}, HMGCN \cite{jin2019fine}) and general etype recognition methods (Sleeman et. al. \cite{sleeman2015entity}, Giunchiglia et. al. \cite{giunchiglia2020entity}). 

\begin{table*}[!t]
\centering
\caption{Quantitative results for our proposed etype recognition algorithm on different datasets. The best and second-best results are highlighted in {\color[HTML]{FF0000} red} and {\color[HTML]{0070C0} blue} colors, respectively.}
  \label{tab1}
\resizebox{1\columnwidth}{!}{

\begin{tabular}{@{}lcccccccc@{}}
\toprule
\multirow{2}*{\textbf{Methods}} & \multicolumn{2}{c}{\textbf{CONF}} & \multicolumn{2}{c}{\textbf{BIBLIO}} & 
\multicolumn{2}{c}{\textbf{EnType$_{Self}$}} & \multicolumn{2}{c}{\textbf{EnType$_{Gen}$}}    \\ \cmidrule(l){2-3}  \cmidrule(l){4-5}  \cmidrule(l){6-7} \cmidrule(l){8-9} & \textbf{ Ma-F1 } & \textbf{ Mi-F1} & \textbf{ Ma-F1 } & \textbf{ Mi-F1} & \textbf{ Ma-F1 } & \textbf{ Mi-F1}  & \textbf{ Ma-F1 } & \textbf{ Mi-F1} \\ \midrule 

FCAMap \cite{chen2019identifying}  & 0.651  & 0.660  & 0.824  & 0.801  
& -   & -     & 0.468   & 0.478          \\
AML \cite{faria2013agreementmakerlight}    & {\color[HTML]{0070C0}  0.717}  &{\color[HTML]{0070C0} 0.723} & 0.845  & 0.831  
& -   & -       & 0.590   & 0.584         \\
LogMap \cite{jimenez2011logmap}  & 0.680  & 0.682   & {\color[HTML]{FF0000} 0.851}  & {\color[HTML]{0070C0} 0.848}
& -  & -   & 0.565   & 0.561   \\
Bento et. al. \cite{bento2020ontology} & 0.708 & 0.714 & 0.806  & 0.816 
& 0.479   & 0.502    & 0.495   & 0.491   \\ 
Nkisi-Orji et. al. \cite{nkisi2018ontology}  & 0.643  & 0.648  & 0.760  & 0.775  
& 0.533  & 0.557   & 0.544   & 0.537   \\

ConnectE \cite{zhao2020connecting} & 0.509  & 0.517 & 0.615  & 0.622
&{\color[HTML]{FF0000} 0.849}  & 0.833 & {\color[HTML]{0070C0} 0.633}  & 0.627 \\


HMGCN \cite{jin2019fine} & 0.487  & 0.511 & 0.590  & 0.588 
& 0.827  & 0.830 & 0.608  & 0.611 \\

Sleeman et. al. \cite{sleeman2015entity} & 0.542 & 0.545 & 0.718 & 0.714 
& 0.759 & 0.770 & 0.430 & 0.458\\
Giunchiglia et. al. \cite{giunchiglia2020entity} & 0.580 & 0.593 & 0.679 & 0.683 
& 0.762 & 0.758 & 0.610 & 0.597\\ 

\hline
ETR$_{XGBoost}$      & {\color[HTML]{FF0000} 0.744} & {\color[HTML]{FF0000} 0.743}   & {\color[HTML]{0070C0} 0.847}   &  {\color[HTML]{FF0000} 0.853}   & {\color[HTML]{0070C0} 0.839}  & {\color[HTML]{FF0000} 0.842}   & 0.628  &{\color[HTML]{0070C0} 0.629}     \\
ETR$_{ANN}$    & 0.693   & 0.689  & 0.828  &  0.840
& 0.823  & {\color[HTML]{0070C0} 0.836} & {\color[HTML]{FF0000} 0.641}   & {\color[HTML]{FF0000} 0.640}      \\

\bottomrule  

\end{tabular}
}
\end{table*}

Table \ref{tab1} shows the comparison between our approach with state-of-the-art methods. Firstly, we can find that our methods ETR$_{XGBoost}$ and ETR$_{ANN}$ achieve promising results, each leading performance on several measurements. To be more specific, ETR$_{XGBoost}$ achieves the best or second-best results of all metrics on datasets CONF and BIBLIO. AML achieves the second-best performance on CONF; LogMap obtains the best and the second-best results of two metrics on BIBLIO, respectively. We find that ontology matching and general etype recognition methods tend to perform better than entity typing methods on datasets CONF and BIBLIO since ConnectE and HMGCN are not designed for schema-level cases. Notice that we use Ma-F1 and Mi-F1 to reflect the performance of classes and samples. Thus, methods have greater Ma-F1 than Mi-F1 (e.g., FCAMap and AML for BIBLIO) means these methods are performing better on few-sample classes and lose in rich-sample classes, which points to an unbalanced result. As for instance-level etype recognition datasets EnType$_{Self}$ and EnType$_{Gen}$, ETR$_{ANN}$ leads the results on EnType$_{Gen}$, and ETR$_{XGBoost}$ achieves the best and second-best results of two metrics on EnType$_{Self}$, respectively. Ontology matching methods poorly perform in instance-level cases since most of them rely on label matching, where the label of entity is commonly not relevant to its etype in practice\footnote{Notice that we remark with `-' where the methods are not applicable to the dataset.}. Deep learning-based entity typing methods also achieve promising performance on EnType$_{Self}$, where ConnectE leads the Ma-F1 score. However, their performance decreases on EnType$_{Gen}$ since they are designed for knowledge completion which utilizes self-information rather than additional ontology information. Thus, datasets using over one ontology are challenging for such methods. 
Etype recognition methods by Sleeman et. al. \cite{sleeman2015entity} and Giunchiglia et. al. \cite{giunchiglia2020entity} only achieve limited results on all four datasets, since they do not distinguish the property information when recognizing the etypes. 
Considering the average results of our approach with different models are performing better or close to the state-of-the-art, we can say that our approach surpasses the state-of-the-art competitors on etype recognition tasks\footnote{All methods do not have significant differences in running times.}. Meanwhile, although ETR$_{XGBoost}$ and ETR$_{ANN}$ produce slightly different results within different ML models, the stable overall performance produced by two different models indicates that our proposed similarity metrics and etype recognition algorithm are valid and adaptive.

\begin{table*}[!t]
\centering
\caption{Quantitative evaluation of instance-level recognition on different entity resolutions.}
\label{tab:5}
\resizebox{1\columnwidth}{!}{

\begin{tabular}{@{}lcccccccc@{}}
\toprule
\multirow{2}{*}{\textbf{Methods}} & \multicolumn{4}{c}{\textbf{Person}} & \multicolumn{4}{c}{\textbf{Organization}}   \\ \cmidrule(l){2-5}  \cmidrule(l){6-9} & \multicolumn{1}{l}{MilitaryPerson} & \multicolumn{1}{l}{Athlete} & \multicolumn{1}{l}{Artist} & \multicolumn{1}{l}{Comedian} & \multicolumn{1}{l}{MilitaryOrg.} & \multicolumn{1}{l}{Company} & \multicolumn{1}{l}{SportsClub} & \multicolumn{1}{l}{ReligiousOrg.} \\ \midrule

ConnectE \cite{zhao2020connecting} & 0.501 & \textbf{0.756}   & 0.679  & 0.640
& 0.627  & \textbf{0.658} & 0.605  & 0.525 \\

HMGCN \cite{jin2019fine} & 0.464  & 0.592  & 0.660  & 0.598 
& 0.643  & 0.507 & 0.680  & 0.531 \\

Sleeman et. al.  \cite{sleeman2015entity} & 0.472 & 0.639  & 0.690 & 0.597 
& 0.479 & 0.463 & 0.637 & 0.510 \\

Giunchiglia et. al. \cite{giunchiglia2020entity} & 0.429 & 0.658   & 0.636 & 0.522 
& 0.512 & 0.487 & 0.624 & 0.482 \\  

ETR$_{XGBoost}$   & 0.507 & 0.722    & \textbf{0.683}  & \textbf{0.659}  
& \textbf{0.655}    & 0.528          & \textbf{0.689}   & 0.491  \\
ETR$_{ANN}$    & \textbf{0.508}  & 0.712 & 0.677  & 0.581   
& 0.627   & 0.487          & 0.663  & \textbf{0.550}\\ 

\bottomrule
\end{tabular}
}
\end{table*}

Considering that etype recognition performance is affected by entity resolutions, we implement an additional experiment based on more specific etypes. We select four sub-classes of etype \textit{person} and \textit{organization} and their corresponding entities as candidate pairs, respectively. We exploit the same ML models as we used in Table \ref{tab1} to compare with existing methods in this experiment. Table \ref{tab:5} presents the Mi-F1 score of the recognition results. We still find our methods achieve better recognition performance than state-of-the-art methods in most cases. ETR$_{XGBoost}$ obtains promising overall performance on specific etype recognition, where entities of \textit{artist} and \textit{comedian} obtain better results in the \textit{person} group and entities of \textit{military unit} and \textit{sports club} perform better in the \textit{organization} group. The experimental results show that our metrics and approach can also be applied for specific instance-level etype recognition.

\subsection{Ablation study}

\noindent \textbf{Effect of similarity metrics.}
The first ablation study is to evaluate if each of the proposed property-based similarity metrics is effective. In this experiment, we test the backbones\footnote{trained by all three metrics ($Sim_V, Sim_H, Sim_I$)} (B) which were used in the etype recognition tasks. Based on the backbones, we also design a controlled group that includes models trained without one of the property-based similarity metrics (i.e. B-Sim$_V$, B-Sim$_H$ and B-Sim$_I$) and models trained without all metrics (i.e. B-L). If the backbones perform better than the corresponding models in the controlled group, we can quantitatively conclude that each of the property-based similarity metrics ($Sim_V, Sim_H, Sim_I$) contributes to the etype alignment and recognition tasks. Table \ref{tab:6} demonstrates the Mi-F1 score of each group, where we apply dataset CONF and EnType$_{Gen}$. Note that we select two models for both cases as Table \ref{tab:6} shows. We find that backbones perform better than models in the controlled group, especially for models trained without all metrics. Thus, we consider all property-based similarity metrics contribute to better recognition performance. Particularly, Sim$_V$ and Sim$_H$ significantly affect the performance of schema-level cases, and Sim$_I$ affects instance-level cases more.

\begin{table}[h]
\centering
\caption{Ablation study on property-based similarity metrics.}
\label{tab:6}
\resizebox{0.8\columnwidth}{!}{
\begin{tabular}{@{}llccccc@{}}
\toprule
\multicolumn{1}{c}{Dataset} & \multicolumn{1}{c}{Model}  & Backbone & B-Sim$_V$ &B-Sim$_H$ & B-Sim$_I$   & B-L\\ \midrule
\multirow{2}{*}{ConfTrack} & ETA$_{ANN}$  & \textbf{0.713}    & 0.635     & 0.639    & 0.660  & 0.618     \\
\addlinespace & ETA$_{XGBoost}$                & \textbf{0.740}    & 0.648     & 0.655    & 0.694  & 0.632     \\ \hline
\multirow{2}{*}{EnType$_{Gen}$} & ETR$_{ANN}$ & \textbf{0.537}  & 0.327  & 0.391  & 0.309  & -    \\
\addlinespace & ETR$_{XGBoost}$                    & \textbf{0.559}  & 0.413  & 0.402  & 0.385  & -     \\\bottomrule
\end{tabular}
}
\end{table}

\noindent \textbf{Effect of constraint factor.}
In section 3.1, we defined a constraint factor $\lambda$ for calculating the metric $Sim_H$. This study aims to statistically identify the value of $\lambda$. We apply the dataset CONF and its two best-performed models. The value of $\lambda$ is set evenly from 0.1 to 1 by discrete points. We evaluate if this pre-set factor affects the final recognition performance and obtain the best value of $\lambda$ for generic etype recognition. Table \ref{tab:7} demonstrates the results, where we highlight both the best and second-best results. We can find that different values of $\lambda$ do affect the final etype recognition performance. And two models show a similar trend that the best value of $\lambda$ is close to 0.5. As a result, we assign $\lambda = 0.5$ to calculate metric $Sim_H$ in our experiments.

\begin{table}[h]
\centering
\caption{Ablation study on the constraint factor $\lambda$.The best and second-best results are highlighted in {\color[HTML]{FF0000} red} and {\color[HTML]{0070C0} blue}, respectively.}
\label{tab:7}
\resizebox{0.99\columnwidth}{!}{
\begin{tabular}{@{}clccccccccc@{}}
\toprule
\multicolumn{1}{c}{Factor} & \multicolumn{1}{c}{Model} & 0.1 & 0.2 & 0.3 & 0.4 & 0.5 & 0.6 & 0.7 & 0.8 & 0.9 \\ \midrule
\multirow{2}{*}{$\lambda$} & ETA$_{ANN}$ 
& 0.613 & 0.638     & 0.670   & 0.678  & {\color[HTML]{FF0000} 0.712} & {\color[HTML]{0070C0} 0.685}  & 0.679 & 0.653  & 0.630 \\
\addlinespace & ETA$_{XGBoost}$   
& 0.662 & 0.677 & 0.714   & {\color[HTML]{0070C0} 0.727} &  {\color[HTML]{FF0000} 0.729} & 0.716 & 0.654 & 0.711 &  0.705  \\ \bottomrule
\end{tabular}
}
\end{table}

\section{Related Work}
\label{sec7}
\subsection{Ontology and schema alignment}
In the early phases, this research focused on string-based methods, including string-based metrics (N-gram, Levenshtein, etc.), syntactic operations (lemmatization, stop word removal, etc.), and semantic analysis (synonyms, antonyms, etc.) \cite{cheatham2013string}. Sun et al, \cite{sun2015comparative} review a wide range of string similarity metrics and propose the ontology alignment method by selecting similarity metrics in different scales. Although string-based methods can lead to effective performance in some cases, selecting the right metric for matching specific datasets is the most challenging part. The structure of an ontology has also been considered important information for identifying etypes \cite{giunchiglia2012s}. Such studies suppose that two etypes are more likely to be aligned if they have the same super-class or sub-class. The LogMap system \cite{jimenez2011logmap} uses a two-step matching strategy, that is, matches two etypes $E_a$ and $E_b$ by a lexical matcher, and then considers the etypes that are semantically close to $E_a$ are more likely to be semantically close to $E_b$. AML \cite{faria2013agreementmakerlight} introduces an ontology matching system that consists of a string-based matcher and a structure-based matcher, building internal correspondences by exploiting \textit{is-a} and \textit{part-of} relationships. Machine learning techniques are also applied to this topic. Some studies model the etype matching task as a classification task, trying to encode the information like string and structure similarities as features for model training. Amrouch et al, \cite{amrouch2016decision} develop a decision tree model by exploiting lexical and semantic similarities of the etype labels to match schemas. 

\subsection{Entity type recognition}
According to the different usage and motivation, studies on entity type recognition (also called entity typing) focus on three main directions: (1) recognizing the type of \textit{entity from text}; (2) recognizing the type of entities from the single ontology for \textit{knowledge completion}; (3) recognizing the type of entities from different ontologies for \textit{ontology merging}. Recent studies focus more on the first two tasks, e.g., \cite{jin2019fine} introduce a hierarchical multi-graph convolutional network for fine-grained entity typing and \cite{zhao2020connecting} propose a deep learning-based entity typing method ConnectE which exploits knowledge embedding features for knowledge completion. Large language models are also introduced for similar tasks, e.g., some Bert-based methods. For instance, \cite{ding2021prompt} exploit BERT-based language models to predict the entity type of words in sentences by inputting prompts. Different from the former two tasks, we focus on recognizing the type of etypes/entities from other ontologies for extending the reference ontology automatically. Rather than using label-based methods, some previous studies also consider properties as a possible solution,  \cite{sleeman2015entity} propose an etype recognition method by modeling etype recognition as a multi-class classification task. However, a pre-filtering step is needed since only properties shared across all candidates are counted for training and testing, which means there will be a few properties remaining after such filtering and a large amount of critical information will be discarded. Thus, the adaptation of such methods will be limited when applied in practice. Giunchiglia and Fumagalli \cite{giunchiglia2020entity} propose a set of metrics for selecting the reference ontology to improve the above method, which achieves improved performance with the support of a large number of ontologies. However, there are still limitations since these studies consider all properties with the same weight and neglect to distinguish properties that will contribute differently during etype recognition.

\section{Conclusions}
In this paper, we have proposed a generic etype recognition algorithm via a set of novel property-based similarity metrics. Firstly, we discuss that the corresponding properties are used to implicitly describe etypes, which provides us with a novel insight for identifying etypes. Then we propose three metrics for measuring the contextual similarity between reference etypes and candidate etypes/entities, namely the horizontal similarity $Sim_H$, the vertical similarity $Sim_V$, and the informational similarity $Sim_I$. Based on our proposed metrics, we develop an ML-based algorithm for etype recognition. Thus, we validate our algorithm for the corresponding data level. Compared with the state-of-the-art methods, the experimental results show the validity of the similarity metrics and the superiority of the proposed etype recognition algorithm, both quantitatively and qualitatively. Our future work will continually focus on fine-tuning the property-based similarity metrics, trying to apply our etype recognition method to more specific ontologies.

\section*{Acknowledgements}

The research conducted by Fausto Giunchiglia 
has received funding from the \emph{InteropEHRate} project, co-funded by the European Union (EU) Horizon 2020 program under grant number 826106, and the research conducted by Daqian Shi has received funding from the program of China Scholarships Council (No.  202007820024).

\bibliographystyle{vancouver}
\bibliography{ICSC2022}

\end{document}